\documentclass[english,journal]{elsarticle}
\usepackage[T1]{fontenc}
\usepackage[latin9]{inputenc}
\usepackage{float}
\usepackage{amsmath}
\usepackage{amssymb}
\usepackage{graphicx}
\usepackage{setspace}
\PassOptionsToPackage{normalem}{ulem}
\usepackage{ulem}
\makeatletter

\providecommand{\tabularnewline}{\\}
\floatstyle{ruled}
\newfloat{algorithm}{tbp}{loa}
\providecommand{\algorithmname}{Algorithm}
\floatname{algorithm}{\protect\algorithmname}

\makeatother

\usepackage{babel}
\begin{document}

\begin{frontmatter}{}

\title{Fisher Information based Stochastic Gradient Ascent for Online Learning
of Dirichlet Process Mixture and Theory.}

\author{Kart-Leong Lim, Xudong Jiang}

\address{Nanyang Technological University, 50 Nanyang Ave, 639798, Singapore }
\begin{abstract}
Scalable algorithms of posterior approximation allow Bayesian nonparametrics
such as Dirichlet process mixture to scale up to larger dataset at
fractional cost. Recent algorithms, notably the stochastic variational
inference performs local learning from minibatch. The main problem
with stochastic variational inference is that it relies on closed
form solution. Stochastic gradient ascent is a modern approach to
machine learning and is widely deployed in the training of deep neural
networks. In this work, we explore using stochastic gradient ascent
as a fast algorithm for the posterior approximation of stick-breaking
Dirichlet process mixture. However, stochastic gradient ascent alone
is not optimal for learning. In order to achieve both speed and performance,
we turn our focus to stepsize optimization in stochastic gradient
ascent. As as intermediate approach, we first optimize stepsize using
the momentum method. Finally, we introduce Fisher information to allow
adaptive stepsize in our posterior approximation. In the experiments,
we justify that our approach using stochastic gradient ascent do not
sacrifice performance for speed when compared to closed form coordinate
ascent learning on these datasets. Lastly, our approach is also compatible
with deep ConvNet features as well as scalable to large class datasets
such as Caltech256 and SUN397. 
\end{abstract}
\begin{keyword}
Dirichlet Process Mixture, Stochastic Gradient Ascent, Fisher Information,
Scalable Algorithm 

\cortext[cor1]{Corresponding author: email: lkartl@yahoo.com.sg}
\end{keyword}

\end{frontmatter}{}

\section{Introduction}

A common Bayesian nonparametrics (BNP) task known as model selection
jointly perform clustering and also estimate the number of clusters
to represent a dataset. It is deployed in many areas such as classification
\cite{zhang2018infinite,wu2018semi} and unsupervised learning \cite{ye2019multi,ma2014bayesian,fan2013variational,liu2019bayesian}.
A well known BNP model is the Dirichlet process mixture (DPM). The
variational inference (VI) of DPM \cite{Blei2006} is mainly based
on the closed form coordinate ascent where it iteratively repeating
its computational task (or algorithm) on the entire sequence of dataset
samples, also known as a batch. The main problems with VI \cite{Blei2006}
are:
\begin{quote}
a) Not scalable.

b) Require closed form solution.
\end{quote}
Scalable algorithms of VI allows BNP to scale up to larger dataset
at fractional cost. The more recent work on scalable BNP in general
is the stochastic variational inference (SVI) \cite{Hoffman2013}.
It is based on stochastic optimization where it performs diminishing
stepsize updates using minibatch. This allows the algorithm to ``see''
the entire dataset especially large datasets when sufficient iterations
has passed. SVI was demonstrated on Bayesian posteriors approximation
and on large scale datasets at 3.8M documents and 300 topics. However,
SVI rely on closed form coordinate ascent as noisy local estimate.
Also, the stepsize is not adaptive. This means that SVI cannot adapt
its stepsize value according to the statistics of the minibatch randomly
drawn each iteration. The main problems with SVI are:
\begin{quote}
c) Require closed form solution.

d) Stepsize not adaptive.
\end{quote}
VI and stochastic gradient ascent (SGA) are mutually exclusive. Recently,
SGA was proposed for VI in \cite{mandt2017stochastic}. Unlike SVI,
SGA do not require closed form expressions. In a stochastic or noisy
setting, SGA alone is inefficient. The main problem with SGA \cite{mandt2017stochastic}
is:
\begin{quote}
e) No stepsize optimization. 
\end{quote}
In our approach to DPM, we seek to address the above problems (a,b,c,d,e)
by optimizing SGA using adaptive stepsize. To achieve this we first
reformula VI as a SGA problem. Next, we use Fisher information to
search along the curvature of the posterior to automatically find
the optimal stepsize. Our approach also ensure that we can achieve
on-par performance with VI and is scalable to large dataset such as
Caltech256 and SUN397. It is rare to find DPM carried out dataset
such as SUN397. The experiments also confirmed that we can handle
large dimensional features such as deep ConvNet features. The novel
contributions in this work are:
\begin{verse}
i) VI using SGA.

ii) SGA using adaptive stepsize.

iii) Scalable to large datasets.
\end{verse}
We also discuss on the conditions for the convergence of SGA, namely
convexity and stepsize optimization, the motiviation behind using
Fisher information, the practical aspect of model selection, our assumptions
for DPM, as well as our objective for solving image classification
using DPM. Lastly, we compare our approach to DPM with state-of-the-arts
methods on six datasets.

\subsection{Related Works}

The use of stepsize to perform VI is certainly not new. We call these
works as Monte Carlo VI (MC-VI) \cite{ranganath2014black,paisley2012variational,kingma2014stochastic,welling2011bayesian,rezende2015variational}.
The key idea of MC-VI is to take the gradient of the evidence lower
bound (ELBO) with a stepsize for updating corresponding VI parameters.
Some notable works in this area include the black box VI \cite{ranganath2014black},
VI with stochastic search \cite{paisley2012variational} and the stochastic
gradient variational Bayes \cite{kingma2014stochastic}. An emphasis
in these works (e.g. black box VI) is the use of Monte Carlo integration
to approximate the expectation of the ELBO. There are two advantages.
Firstly, there is no need to constrain the learning of variational
posteriors expectation to analytical solution. Secondly, Monte Carlo
approximated solutions lead to true posteriors at the expense of greater
computational cost. However, most works usually demonstrated MC-VI
on non-DPM related task such as the logistic regression \cite{welling2011bayesian,paisley2012variational}.
Non-DPMs do not deal with incomplete data or cluster label issue.
Thus, the learning of the MC-VI \cite{ranganath2014black,paisley2012variational,kingma2014stochastic,welling2011bayesian}
described above are more suitable to relatively simpler parameter
inference problems. Also, these works on MC-VI were mainly demonstrated
on datasets with smaller datasets of 10 classes with 60K image sizes
such as MNIST. In this paper we target BNP task of up to 108K images
with 397 object classes and high dimensions features of 4096 using
VGG16 pretrained on ImageNet and Place205. 

A different technique known as SGA for VI (SGA-VI) was concurrently
proposed in \cite{mandt2017stochastic}, it was mainly demonstrated
on the logistic regression model for classification. Our learner for
DPM also falls under the SGA-VI category. Instead of Monte Carlo integration,
SGA-VI methods use Maximum A Posterior (MAP) to approximate posteriors
(or the expectation of the ELBO) for simplicity. In the pursuit of
optimizing SGA, two broad categories exist. Either diminishing the
stepsize \cite{robbins1985stochastic} (this is to avoid SGA bouncing
around the optimum of the objective function) or automatically finding
the optimal stepsize \cite{tan2016barzilai}. In the SGA-VI approach
in \cite{mandt2017stochastic}, the authors mainly focus on constant
stepsize. Most MC-VI \cite{welling2011bayesian,paisley2012variational}
also only use constant stepsize. Our empirical finding suggest that
constant stepsize for SGA-VI is not ideal since convergence is slow.
Instead, we use Fisher information to automatically adapt the stepsize
in SGA-VI. While the use of Fisher information has already been proposed
to VI \cite{Hoffman2013,honkela2010approximate} and non VI algorithms
\cite{duchi2011adaptive,kingma2014adam}, to our best knowledge this
is the first time it has been applied to SGA-VI. In \cite{honkela2010approximate},
the authors applied Fisher information to DPM using VI. In SVI \cite{Hoffman2013},
the Fisher information manifest in diminishing stepsize. The rate
of stepsize decay is fixed. In our SGA-VI approach, the Fisher information
manifest in adaptive stepsize. The stepsize is automatically found.

\subsection{What is Online and Offline?}

\uline{DPM:} In the context of BNP such as DPM, there is a distinction
of online vs offline DPM. The online DPM originates from the Chinese
restaurant process while the offline DPM is represented using a Stick-breaking
process. The online approach \cite{Kulis2012,broderick2013streaming}
starts with a single cluster and for each new sample, it makes a decision
whether or not to introduce a new cluster. On the other hand, the
offline approach \cite{Blei2006,Kurihara2009} starts with a near
infinite \# of clusters. For each step towards convergence, it prune
off ineffective clusters and finds the optimal \# of clusters to represent
the entire dataset. 

\uline{VI:} In the context of VI, the online approach can either
refer to ones that use sequential processes of VI to adapt to streaming
data \cite{broderick2013streaming,sato2001online,fan2016online,Hoffman2013}.
The offline approach mainly refers to traditional VI that runs on
the entire dataset. The closed form coordinate ascent \cite{Blei2006,fan2013variational,ma2014bayesian}
is such a case.

\uline{SGA:} Yet in the optimization of stochastic gradient, the
notion of online vs offline learning would totally differ from the
above two contexts. The well known online learners are Adam \cite{kingma2014adam},
Adagrad \cite{duchi2011adaptive,patacchiola2017head}, RMSprop and
etc. These techniques are rarely found in BNP or VI setting. The offline
approach in stochastic gradient refers to batch gradient ascent where
the entire training dataset is used.

\uline{VAE:} Lastly, VI in deep generative models \cite{kingma2014stochastic,li2018discriminatively}
is not to be confused with VI in BNP. The former is associated with
the gradient update of network weights \cite{kingma2014stochastic}
and the latter (this work) is concerned with the posterior approximation
of hidden variables.

Of the above, our work falls under the categories of offline DPM,
online VI and online SGA.

\section{Background: Variational Inference and the Dirichlet Process Mixture}

We assume the mixture model in DPM is Gausian distributed. DPM models
a set of $N$ observed variable denoted as $x=\left\{ x_{n}\right\} _{n=1}^{N}\in\mathbb{R}^{D}$
with a set of hidden variables, $\theta=\left\{ \mu,z,v\right\} $.
The total dimension of each observed instance is denoted $D$. The
mean is denoted $\mu=\left\{ \mu{}_{k}\right\} _{k=1}^{K}\in\mathbb{R}{}^{D}$.
We assume diagonal covariance i.e. $\varSigma_{k}=\sigma_{k}^{2}I$
and define variance as a constant, $\sigma_{k}=\sigma\in\mathbb{R}{}^{D}$.
The cluster assignment is denoted $z=\left\{ z_{n}\right\} _{n=1}^{N}$
where $z_{n}$ is a $1-of-K$ binary vector, subjected to $\sum_{k=1}^{K}z_{nk}=1$
and $z_{nk}\in\left\{ 0,1\right\} $. We follow the truncated stick-breaking
\cite{Blei2006} process to DPM with truncation level denoted $K$.
Starting with a unit length stick, the proportional length of each
broken off piece (w.r.t remainder of stick) is a random variable $v$
drawn from a Beta distribution and is denoted $v=\left\{ v_{k}\right\} _{k=1}^{K}\in\mathbb{R}$.
All broken off pieces add up to a unit length of the full stick. This
value can be seen as the the cluster weight and is given as $\pi_{k}=v_{k}\prod_{l=1}^{k-1}\left(1-v_{k}\right)$. 

In the Bayesian approach to DPM, each hidden variable is now modeled
by a distribution as follows

\begin{equation}
\begin{array}{c}
x\mid z,\mu\sim\mathcal{N}\left(\mu,\sigma\right){}^{z}\\
z\mid v\sim Mult(\pi)\\
\mu\sim\mathcal{N}\left(m_{0},\lambda_{0}\right)\\
v\sim Beta(1,a_{0})
\end{array}
\end{equation}

We refer to $\mathcal{N},Beta,Mult$ as the Gaussian, Beta and Multinomial
distribution respectively. The terms $\lambda_{0}$ and $m_{0}$ refer
to the Gaussian prior hyperparameters for cluster mean. $a_{0}$ is
the Beta prior hyperparameter. The hyperparameters are treated as
constants. 

\subsection{Variational Inference and Learning}

Variational inference \cite{bishop2006pattern} approximates the intractable
integral of the marginal distribution$\begin{array}{c}
p(x)=\int p(x\mid\theta)p(\theta)d\theta\end{array}$. This approximation can be decomposed as a sum, $\ln p(x)=\mathcal{L}+KL_{divergence}$
where $\mathcal{L}$ is a lower bound on the joint distribution between
observed and hidden variable. A tractable distribution $q(\theta)$
is used to compute $\mathcal{L}=\int q(\theta)\ln\left(\frac{p(x,\theta)}{q(\theta)}\right)d\theta$
and $KL_{divergence}=-\int q(\theta)\ln\left(\frac{p(\theta|x)}{q(\theta)}\right)d\theta$.
When $q(\theta)=p(\theta|x)$, the Kullback-Leibler divergence is
removed and $\mathcal{L}=\ln p(X)$. The tractable distribution $q(\theta)$
is also called the variational posterior distribution and assumes
the following factorization $q(\theta)=\prod_{i}q(\theta_{i})$. Variational
log posterior distribution is expressed as $\ln q(\theta_{j})=E_{i\neq j}\left[\ln p(x,\theta_{i})\right]+const.$

The DPM joint probability can be derived as follows
\begin{equation}
\begin{array}{c}
p(x,\mu,z,v)=p(x\mid\mu,z)p(\mu)p(z\mid v)p(v)\end{array}
\end{equation}
Next, we treat the hidden variables as variational log posteriors,
$\ln q\left(\theta\right)$ and we use mean-field assumption and logarithm
to simplify their expressions \cite{bishop2006pattern}
\begin{equation}
\ln q(\mu,z,v)=\ln q(\mu)+\ln q(z)+\ln q(v)
\end{equation}
the variational log-posteriors are defined using expectation functions
\cite{bishop2006pattern} as follows 
\begin{equation}
\begin{array}{c}
\ln q(\mu)=E_{z}\left[\ln p(x\mid\mu,z)+\ln p(\mu)\right]+const.\\
\ln q(v)=E_{z}\left[\ln p(z\mid v)+\ln p(v)\right]+const.\\
\ln q(z)=E_{\mu}\left[\ln p(x\mid\mu,z)+\ln p(z\mid v)\right]+const.
\end{array}
\end{equation}

Variational log-posteriors such as the ones in eqn (4) are analytically
intractable to learn. Welling and Kurihara \cite{Kurihara2009} discussed
a family of alternating learners and they are classified into Expectation
Expectation (EE), Expectation Maximization (EM), Maximization Expectation
(ME) and Maximization Maximization (MM). Our main interest is the
MM algorithm due to its simplicity as demonstrated in \cite{lim2018fast}.
In the learning of VI or closed form coordinate ascent algorithm,
the goal is to alternatively compute the expectations of $\ln q\left[\theta\right]$
and $\ln q\left[z\right]$ (defined as $E\left[\theta\right]$ and
$E\left[z\right]$ respectively) each iteration. In DPM, $\theta$
refers to $\mu$ and $v$. In MM, these expectations are actually
MAP estimated and the objective function of MM is given as

\begin{equation}
\begin{array}{c}
E\left[z\right]\leftrightarrow E\left[\theta\right]\\
\approx\;\;\underset{z}{\arg\max}\;\ln q(z)\leftrightarrow\underset{\theta}{\arg\max}\;\ln q(\theta)
\end{array}
\end{equation}

The argument for the convergence of MM is briefly discussed as follows
\cite{lim2018fast}:
\begin{quote}
``By restricting the mixture component to exponential family, the
RHS is a convex function, hence a unique maximum exist. In the LHS,
for independent data points $x_{1},...x_{N}$ with corresponding cluster
assignment $z_{1},...,z_{N}$ , given sufficient statistics each data
point should only fall under one cluster, meaning that for LHS there
is indeed a unique maximum. It is further discussed in \cite{Neal1998,Titterington2011a,bishop2006pattern}
that the above function monotonic increases for each iteration and
will lead to convergence.''
\end{quote}

\section{Proposed SGA Learner for Variational Posteriors}

Previously in VI, variational log posteriors are updated by computing
their expectation via closed form solution. In this section, we wish
to estimate these expectations (of variational log posteriors) using
SGA. We start with the constant stepsize SGA, then we present two
more variants which introduce decaying stepsize to SGA. 

\subsection{Stochastic gradient ascent with constant stepsize (SGA)}

We propose how to perform learning on a variational log posterior,
$\ln q\left(\theta\right)$ using SGA. Recall that previously, the
goal of VI is to find an expression for estimating the expectation
of $\ln q\left(\theta\right)$ i.e. $E\left[\theta\right]$. Instead
of estimating a closed form expression for $E\left[\theta\right]$
in VI, using the MAP approach which is a global maximum, we now seek
the local maximum of $\ln q\left(\theta\right)$ as below
\begin{equation}
\begin{array}{c}
E\left[\theta\right]=\arg\max_{\theta}\;\ln q\left(\theta\right)\\
=E\left[\theta\right]^{'}+\eta\nabla_{\theta}\ln q(\theta)
\end{array}
\end{equation}

$E\left[\theta\right]^{'}$ refers to the initial or previous iteration
value of $E\left[\theta\right]$ and $\eta$ is the learning rate.

We can ensure the convergence of $E\left[\theta\right]$ using SGA
by i) convexity checking and ii) value of SGA stepsize.
\begin{verse}
i) Convexity test: When the condition $\nabla_{\theta}^{2}\ln q(\theta)\leq0$
is fulfilled, the function is concave and vice-versa. 

ii) For a function $f$ that is differentiable and convex and its
gradient is Lipschitz continuous, SGA will converge to a local maxima
when it satisfies the condition \cite{tan2016barzilai} at current
iteration $t$, 
\begin{equation}
\eta_{t}=\frac{\left\Vert s_{t}\right\Vert _{2}^{2}}{s_{t}^{T}y_{t}}
\end{equation}
where $s_{t}=x_{t}-x_{t-1}$ and $y_{t}=\nabla f(x_{t})-\nabla f(x_{t-1})$.
\end{verse}

\subsection{Stochastic gradient ascent with momentum (SGA+M)}

A standard SGA technique in backpropagation is to apply the momentum
method to smoothen the gradient search as it reaches the local optimum.
The update is computed as

\begin{equation}
\begin{array}{c}
E\left[\theta\right]=E\left[\theta\right]^{'}+\gamma_{new}\\
\gamma_{new}=\alpha\gamma_{old}+\eta\nabla_{\theta}\ln q(\theta)
\end{array}
\end{equation}

The purpose of $\alpha$ in eqn (8) is to slow the step size learning
but it is a constant value and typically fixed at 0.9. When $\alpha=0$,
we recover SGA. For momentum, the computation only involves an additional
memory caching of the computed $\gamma_{old}$.

\subsection{Stochastic gradient ascent with Fisher information (SGA+F)}

Since we are dealing with an approximate posterior or VLP which is
assumed convex, a more superior gradient learning is the natural gradient
learning. It uses Fisher information matrix, $G=E\left[\nabla_{\theta}\ln q(\theta)(\nabla_{\theta}\ln q(\theta))^{T}\right]$,
as the steepest ascent directional search is in Riemannian space.
Natural gradient learning is superior to gradient learning because
the shortest path between two point is not a straight-line but instead
falls along the curvature of the VLP objective \cite{honkela2007natural}.

In order to adapt the stepsize to a stochastic setting (i.e. randomly
drawn minibatch samples) rather than fixing a value (e.g. $\alpha=0.9$
in SGA+M), we can utilize Fisher information $F$ as the direction
of steepest gradient ascent \cite{honkela2007natural} as follows
\begin{equation}
E\left[\theta\right]=E\left[\theta\right]^{'}+F_{\theta}^{-1}\eta\nabla_{\theta}\ln q(\theta)
\end{equation}
Where we define Fisher information as follows
\begin{equation}
\begin{array}{c}
F_{\theta}=E\left[\nabla_{\theta}\ln q(\theta)\circ\nabla_{\theta}\ln q(\theta)\right]\end{array}
\end{equation}
In the above expression, $F_{\theta}$ is a scalar when we assume
diagonal covariance (i.e. each dimension is independent), and $\circ$
refers to element wise product.

There are two main advantages to this technique:
\begin{verse}
i) Low overhead cost: When considering empirical Fisher $F_{\theta}'$
over a set of $M$ random samples, the expectation in eqn (10) becomes
as follows
\begin{equation}
\begin{array}{c}
F_{\theta}'=\frac{1}{M}\sum_{n=1}^{M}\left(\nabla_{\theta}\ln q(\theta)\right)^{2}\end{array}
\end{equation}
There is practically no overhead cost when computing $F_{\theta}'$
since $\nabla_{\theta}\ln q(\theta)$ is a byproduct when computing
eqn (9). 

ii) $F_{\theta}$ is the optimal path: When considering an arbitrary
update $d$ towards the objective in distribution space, we wish to
find the $d$ that maximize our VLP as follows \cite{martens2014new}
\begin{equation}
\begin{array}{c}
d^{*}=\arg\max_{d}\;\ln q\left(\theta+d\right)\\
s.t.\;D_{KL}\left(q(\theta)\parallel q(\theta+d)\right)\approx\frac{1}{2}d^{T}F_{\theta}d
\end{array}
\end{equation}
We further constraint the first row in eqn (12) to a metric based
on KL divergence instead of Euclidean. When we maximize eqn (12) w.r.t
$d$, we can obtain with the following \cite{martens2014new}
\begin{equation}
d\propto-F_{\theta}^{-1}\nabla_{\theta}\ln q(\theta)
\end{equation}
In other words, the optimal update $d$ (in terms of KL divergence)
is determined by gradient $\nabla_{\theta}\ln q(\theta)$ and Fisher
information $F_{\theta}$.
\end{verse}

\section{Unsupervised Image Classification via Proposed SGA Learner}

\subsection{Objective and Motivation using DPM}

In order to demonstrate our new SGA learner, we apply it to model
image classification datasets using DPM. The main reasons are:
\begin{verse}
i) Local estimate vs optimal solution: SGA learners are local estimates
and are usually employed when closed form solutions are not possible.
Fortunately, the inference of DPM is well studied using closed form
solution. Thus, using DPM we have a case-control scenario where we
can access how well SGA learners performed compared to its closed
form counterpart. We quantify this finding in Section 5.

ii) Scalable to large dataset: The closed form solution to DPM i.e.
VB-DPM \cite{Blei2006} suffers from scalability issue. Our SGA learner
avoids this problem by stochastic learning using minibatch and adaptive
stepsize. We compare the CPU time of techniques using closed form
vs our proposed SGA learners in Section 5.
\end{verse}

\subsection{Training DPM using proposed SGA learner}

We propose in Algo. 1, an algorithm for training DPM using SGA. In
DPM, our main goal is to update two continuously distributed variational
posteriors, $\ln q\left(\mu\right)$ and $\ln q\left(v\right)$ each
iteration. $\ln q\left(z\right)$ is a discrete distribution and is
solved traditionally using MAP estimation. For brevity we use the
following expression
\begin{equation}
g\left(\theta\right)=\frac{1}{M}\sum_{n=1}^{M}\nabla_{\theta}\ln q(\theta)
\end{equation}
to represent the gradient averaged from each minibatch for $\ln q\left(\mu\right)$
and $\ln q\left(v\right)$. As seen in Section 3, all SGA discussed
here fundamentally require solving for $\nabla_{\theta}\ln q(\theta)$.

\uline{Gradient of \mbox{$\ln q\left(\mu\right)$}:} In its simplest
SGA form, solving the expectation of $\ln q\left(\mu\right)$ is expressed
as
\begin{equation}
\begin{array}{c}
E\left[\mu\right]=E\left[\mu\right]^{'}+\eta g\left(\mu\right)\end{array}
\end{equation}

The gradient update is expressed as follows
\begin{equation}
\begin{array}{c}
\nabla_{\mu_{k}}\ln q(\mu_{k})=\nabla_{\mu_{k}}E_{z}\left[\ln p(x_{n}\mid\mu_{k},z_{nk})+\ln p(\mu_{k})\right]\\
=\nabla_{\mu_{k}}\left(-\frac{(x_{n}-\mu_{k})^{2}}{2\sigma^{2}}E\left[z_{nk}\right]-\frac{\lambda_{0}(\mu_{k}-m_{0})^{2}}{2\sigma^{2}}\right)\\
=\left(\frac{(x_{n}-\mu_{k})}{\sigma^{2}}E\left[z_{nk}\right]-\frac{\lambda_{0}(\mu_{k}-m_{0})}{\sigma^{2}}\right)
\end{array}
\end{equation}

\uline{Gradient of \mbox{$\ln q\left(v\right)$}:} Next, when defining
the expectation of $\ln q\left(v\right)$ using SGA as follows
\begin{equation}
\begin{array}{c}
E\left[v\right]=E\left[v\right]^{'}+\eta g\left(v\right)\end{array}
\end{equation}

we need to define the gradient below as follows
\begin{equation}
\begin{array}{c}
\nabla_{v_{k}}\ln q(v_{k})=\nabla_{v_{k}}E_{z}\left[\ln p(z\mid v_{k})+\ln p(v_{k})\right]\\
=\nabla_{v_{k}}\;\left\{ E\left[z_{nk}\right]\ln v_{k}\right.\\
\left.+\sum_{j=k+1}^{K}\ln\left(1-v_{k}\right)E\left[z_{nj}\right]+\sum_{k=1}^{K}\left(a_{0}-1\right)\ln\left(1-v_{k}\right)\right\} \\
=\frac{E\left[z_{nk}\right]}{v_{k}}-\frac{\sum_{j=k+1}^{K}E\left[z_{nj}\right]}{1-v_{k}}-\frac{\left(a_{0}-1\right)}{1-v_{k}}
\end{array}
\end{equation}

\uline{MAP learning of \mbox{$\ln q\left(z\right)$}:} Since we
have a fixed number of states and $z_{n}$ is represented by a $1-of-K$
vector, the expectation of $\ln q\left(z\right)$ is obtained by the
MAP estimate of $\ln q(z)$ as follows

\begin{equation}
\begin{array}{c}
E\left[z_{nk}\right]=\arg\max_{z_{nk}}\;\ln q(z_{nk})\\
=\arg\max_{z_{nk}}\;E_{\mu,v}\left[\ln p(x_{n}\mid z_{nk},\mu_{k})+\ln p(z_{nk}\mid v_{k})\right]\\
=\arg\max_{z_{nk}}\;\left\{ \ln E\left[v_{k}\right]+\sum_{l=1}^{k-1}\ln(1-E\left[v_{l}\right])\right.\\
\left.+\ln\left(\frac{1}{\sigma}\right)-\frac{\left(x_{n}-E\left[\mu_{k}\right]\right){}^{2}}{2\sigma^{2}}\right\} z_{nk}
\end{array}
\end{equation}

\subsection{DPM Convexity }

We conduct the convexity checking for our claims over the convergence
of DPM using SGA. When the condition $\nabla_{\theta}^{2}\ln q(\theta)\leq0$
is fulfilled, the function is concave and vice-versa. We show the
conditions for DPM as follows
\begin{equation}
\begin{array}{c}
\nabla_{\mu_{k}}^{2}\ln q(\mu_{k})=-\frac{1}{\sigma^{2}}E\left[z_{nk}\right]-\frac{\lambda_{0}}{\sigma^{2}}\end{array}
\end{equation}
\begin{equation}
\begin{array}{c}
\nabla_{v_{k}}^{2}\ln q(v_{k})=-\frac{E\left[z_{nk}\right]}{v_{k}^{2}}-\frac{\sum_{j=k+1}^{K}E\left[z_{nj}\right]+\left(a_{0}-1\right)}{\left(1-v_{k}\right)^{2}}\end{array}
\end{equation}
In eqn (20) when $E\left[z_{nk}\right]=1$, the condition for convexity
is met. Similarly in eqn (21), when $RHS\leq0$, we have a concave
function. 

\subsection{Assumptions for DPM}

Mixture model such as DPM or GMM which employ diagonal covariance
matrix suffers from a lack of expression for correlations between
each dimensional. However, there are two good reasons. Firstly, they
are scalable to high dimensionality such as deep learning features
since each dimension is treated as independent. Thus, the variational
inference closed form solution or the proposed gradient ascent solution
in this work are less expensive to work with than their multidimensional
counterparts. Secondly, computing the $D\times D$ covariance matrix
on higher dimension is non-trival and is often accomplished using
complex estimators such as probabilistic PCA \cite{bishop2006pattern}
or arbitrary covariance matrix \cite{ari2012maximum}.

\subsection{Performing Model Selection}

We introduce our proposed SGA learner for DPM as shown in Algorithm
1. The learning objective of MM in eqn (5) is to iteratively update
$E\left[\theta\right]\leftrightarrow E\left[z\right]$ until convergence.
It is the same case when using SGA. 

We start off by fixing a truncation level, $K$ that is much larger
than ground truth such that it is independent of the ground truth
(eg. $K=50$ when the ground truth is $<10$, or $K=100$ when the
ground truth is $>30$) as seen in Table 2. 

At the end of each iteration we run the threshold check, $E\left[v_{k}\right]<THR$
for $k=1,...,K$. When $E\left[v_{k}\right]$ does not meet the threshold,
we simply discard this $k^{th}$ cluster and its associated expectations
i.e. $E\left[\mu_{k}\right],E\left[v_{k}\right]$. Thus, we have a
lesser cluster to compute for each case of $E\left[v_{k}\right]<THR$
at the next iteration. We can decide what value of $THR$ to use by
plotting the bar plot of $E\left[v_{k}\right]$ for all $K$ after
a maximum number of iterations has passed. In the early iterations,
we will see there are dominant clusters and some insignificant ones.
However, a typical setting is to use $THR>0$ and let the algorithm
run till it converge. 

Another requirement is to re-arrange $E\left[v_{k}\right]$ according
to descending value. After re-ordering $E\left[v_{k}\right]$, we
must also align $E\left[\mu_{k}\right]$ before computing $E\left[z_{nk}\right]$.
This is due to the representation of the stick-breaking process, $\sum_{l=1}^{k-1}\ln(1-E\left[v_{l}\right])$
appearing in DPM. Ideally, we want to give emphasis to clusters with
larger values of $E\left[v_{k}\right]$ to appear more frequently
rather than random appearance. This is also discussed in \cite{Kurihara2007}.
We found that for most datasets, performing this re-ordering usually
improve the result.

\subsection{Negative lower bound}

Fundamentally, we are solving VI as seen in eqn (5). Except that we
are using SGA as the preferred learner here over closed form coordinate
ascent. Thus, we can check the convergence of learning DPM by checking
that the negative lower bound reaches saturation after sufficient
iterations has passed. The lower bound is computed as follows: 

\begin{equation}
\begin{array}{c}
\mathcal{L}\geq E\left[\ln p(x,z,\mu,v)\right]-E\left[\ln q(z,\mu,v)\right]\\
=E\left[\ln p(x\mid z,\mu)\right]+E\left[\ln p(\mu)\right]+E\left[\ln p(z\mid v)\right]+E\left[\ln p(v)\right]\\
-E\left[\ln q(z)\right]-E\left[\ln q(v)\right]-E\left[\ln q(\mu)\right]
\end{array}
\end{equation}

Each expectation function within the lower bound are computed below
as follows

\begin{equation}
\begin{array}{c}
E\left[\ln p(x\mid z,\mu)\right]=\sum_{k=1}^{K}\sum_{n=1}^{N}-\frac{1}{2}\left(x_{n}-E\left[\mu_{k}\right]\right){}^{2}E\left[z_{nk}\right]\end{array}
\end{equation}

\begin{equation}
\begin{array}{c}
E\left[\ln p(\mu)\right]=\sum_{k=1}^{K}-\frac{\lambda_{0}}{2}(E\left[\mu_{k}\right]-m_{0})^{2}\end{array}
\end{equation}

\begin{equation}
\begin{array}{c}
E\left[\ln p(z\mid v)\right]=\sum_{n=1}^{N}\sum_{k=1}^{K}\ln E\left[v_{k}\right]E\left[z_{nk}\right]\\
+\sum_{j=k+1}^{K}\ln(1-E\left[v_{j}\right])E\left[z_{nj}\right]
\end{array}
\end{equation}

\begin{equation}
E\left[\ln p(v)\right]=\sum_{k=1}^{K}+\left(a_{0}-1\right)\ln\left(1-E\left[v_{k}\right]\right)
\end{equation}

\begin{flushleft}
\begin{equation}
\begin{array}{c}
E\left[\ln q(z)\right]=\sum_{k=1}^{K}\left\{ \sum_{n=1}^{N}-\frac{1}{2}(x_{n}-E\left[\mu_{k}\right])^{2}E\left[z_{nk}\right]\right.\\
\left.+\sum_{n=1}^{N}E\left[z_{nk}\right]\ln E\left[v_{k}\right]+\sum_{j=k+1}^{K}\ln\left(1-E\left[v_{j}\right]\right)E\left[z_{nj}\right]\right\} 
\end{array}
\end{equation}
\par\end{flushleft}

\begin{equation}
\begin{array}{c}
E\left[\ln q(v)\right]=\sum_{k=1}^{K}\left\{ \left(a_{0}-1\right)\ln\left(1-E\left[v_{k}\right]\right)\right.\\
\left.+\sum_{n=1}^{N}E\left[z_{nk}\right]\ln E\left[v_{k}\right]+\sum_{j=k+1}^{K}\ln\left(1-E\left[v_{j}\right]\right)E\left[z_{nj}\right]\right\} 
\end{array}
\end{equation}

\begin{equation}
\begin{array}{c}
E\left[\ln q(\mu)\right]=\sum_{k=1}^{K}\left\{ \sum_{n=1}^{N}-\frac{1}{2}\left(x_{n}-E\left[\mu_{k}\right]\right){}^{2}E\left[z_{nk}\right]\right.\\
\left.-\frac{\lambda_{0}(E\left[\mu_{k}\right]-m_{0})^{2}}{2}\right\} 
\end{array}
\end{equation}

\begin{algorithm}
1) Input: $x\leftarrow\left\{ minibatch\right\} $
\begin{verse}
$\:$
\end{verse}
2) Output: $E\left[\mu_{k}\right],E\left[v_{k}\right],E\left[z_{nk}\right]$
\begin{verse}
$\:$
\end{verse}
3) Initialization:
\begin{verse}
i) $a_{0}=N,\;1e^{-1}\leq\eta\leq1e^{-3}$

ii) $K$ is the user defined trunction level.

iii) use kmeans to initialize $E\left[\mu_{k}\right]$ then compute
$E\left[z_{nk}\right]$ followed by $E\left[v_{k}\right]$

$\:$
\end{verse}
5) Repeat until max iteration or convergence
\begin{verse}
i) Compute $E\left[\mu_{k}\right]$ using SGA

$\:$

ii) Compute $E\left[z_{nk}\right]$ using MAP

$\:$

iii) Compute $E\left[v_{k}\right]$ using SGA

$\:$
\end{verse}
6) Prune $E\left[\mu_{k}\right],E\left[v_{k}\right],E\left[z_{nk}\right]$
for all $E\left[v_{k}\right]<THR$ 
\begin{verse}
\caption{Training DPM using proposed SGA learner}
\end{verse}
\end{algorithm}

\section{Experiment 1: Proposed SGA learner vs traditional VI learner}

Recall in Section 4, our objectives in this experiment is to first
ensure 1) the proposed SGAs should return on-par performance to MM
\cite{lim2018fast}. The minibatch sizes are empirically adjusted
to reflex this. 2) Next, we measure the computational time. The experimental
results should justify our claim that as fast algorithm for computing
DPM, SGAs but must be faster to compute, but must retain on-par performance
to traditional VI learner .

\subsection{Evaluation metric }

In the followings, we evaluate five criterias: 
\begin{verse}
i) Sample size per iteration (Fig. 1)

ii) Computational time (Fig. 1)

iii) NMI (Fig. 2) 

iv) Accuracy (Fig. 2) 

v) Estimated model (Table 2)
\end{verse}

\subsubsection{Sample size per iteration}

Batch learner such as MM typically takes the entire dataset for computation
per iteration. This is also due to the theoretical definition behind
VI. Whereas in stochastic learning, both SGA and SVI runs on a random
subset of samples or minibatch per iteration. When the number of samples
per dataset is too large for batch optimization, we \textbf{limit}
this batch size to a random subset of the entire dataset (e.g. $N=5000)$
as seen in Fig. 1. The main difference between batch and minibatch
learners is that for batch, this subset of samples are fixed for each
iteration. Whereas for minibatch, the subset of samples are not fixed
and each iteration has a random chance to ``see'' the entire dataset.
Empirically, we found that having at least 20 images per class for
defining the minibatch size is necessary for sufficient statistics.

\subsubsection{NMI, ACC, Model}

We use Normalized Mutual Information (NMI) and Accuracy to evaluate
the performance of our learning. For NMI and Accuracy (based on Hungarian
mapping) we use the code in \cite{CHH05}. The definition for Accuracy
(clustering) and NMI is as follows
\[
ACCURACY=\frac{\sum_{n=1}^{N}\delta\left(gt_{n},\:map\left(mo_{n}\right)\right)}{N}
\]
\[
NMI=\frac{MU_{info}(gt,mo)}{\max\left(H\left(gt\right),H\left(mo\right)\right)}
\]
where $gt,mo,map,\delta\left(\cdotp\right),MU_{info},H$ refers to
ground truth label, model's predicted label, permutation mapping function,
delta function ($\delta(gt,mo)=1$ if $gt=mo$ and equal 0 otherwise),
mutual information and entropy respectively. We can see that for Accuracy,
it is measuring how many times the model can produce the same label
as the ground truth on average. For NMI, it is measuring how much
is the overlap or mutual information between ground truth entropy
and model entropy. Model refers to the model selection estimated by
each approach. 

\subsubsection{Computational time}

For a given budget, it is often unnecessary to run DPM until full
convergence especially when the dataset or class is huge. Instead,
we observed that typically in the early stages, DPM aggressively prune
away computed empty clusters $(v_{k}=0)$ from its given initial large
cluster trucation size and the pruning will slowly become relunctant
after around sufficient iterations has passed. This is where it signifies
it has converged to a certain number of dominant clusters close to
ground truth. Empirically, we found that typically 30 to 60 iterations
in our experiments is sufficient for good compromise between performance
and CPU cost.

\subsubsection{Dataset size}

The object and scene categorization datasets used in our experiments
are detailed in Table 1. There are 3 objects and 3 scene datasets
in total. We split the datasets into training or test partition. The
total number of images can range from 3K to 108K. Ground truth refers
to the number of classes per dataset. It ranges from 10 to about 400
classes or clusters in our case. Also, for unsupervised learning we
do not require class labels for learning our models. However, we require
setting a truncation level for each dataset as our model cannot start
with an infinite number of clusters in practice. We typically use
a large truncation value (e.g. $K=1000$ for SUN397) from the ground
truth to demonstrate that our model is not dependent on ground truth
as shown in Table 2. 

For all datasets, we do not use any bounding box information for feature
extraction. Instead, we represent the entire image using a feature
vector. For our DPM input, we mainly extract FC7 of VGG16 as our image
feature extractor. The FC7 feature dimension is 4096. We use VGG16
pretrained on Imagenet. 

\begin{table}
\begin{centering}
\begin{tabular}{|c||c||c|}
\hline 
Dataset & Train & Test\tabularnewline
\hline 
\hline 
Caltech10 & 300 & 2744\tabularnewline
\hline 
\hline 
Caltech101 & 3030 & 5647\tabularnewline
\hline 
\hline 
Caltech256 & 7680 & 22,102\tabularnewline
\hline 
\hline 
Scene15 & 750 & 3735\tabularnewline
\hline 
\hline 
MIT67 & 3350 & 12,270\tabularnewline
\hline 
\hline 
SUN397 & 39,700 & 69,054\tabularnewline
\hline 
\end{tabular}
\par\end{centering}
\caption{Datasets partitioned (object, scene) for Bayesian nonparametrics}
\end{table}

\subsection{Experimental results}

In this section, we visually compare (Fig. 1-2 and Table 2) the 5
evaluation criterias between the MM learner \cite{lim2018fast} with
our newly proposed SGA learners (with learning rate $\eta=0.1$) for
the DPM model on six datasets (Table 1). Specifically, we compare
MM with ``SGA+F'' and ``SGA+M''. We rerun the experiments for
at least 5 times and take their average results for each dataset. 

We first look at dataset \#1 (Caltech10) and \#4 (Scene15). Both are
considered small datasets with less than 5K images and 10-15 classes.
We set the initial truncation level to 50. We ran both MM (full dataset)
and SGA (minibatch of 100 and 300 respectively) for 30 iterations
each. We observed that all SGAs outperforms MM on all 5 evaluation
criteria for both datasets in terms of NMI and Accuracy, closer to
ground truth model selection, faster to compute as a direct consequence
of smaller sample size per iteration. The clustering performance (ACC
\& NMI) of the proposed methods are indeed on-par or better than MM.
Also, it is evident that using SGAs allows it to significantly reduce
the computation time as compared to its batch counterpart MM. This
finding is actually consistent throughout all the datasets.

Next, we look at dataset \#2 (Caltech101), \#5 (MIT67) and \#3 (Caltech256).
These datasets are slightly more challenging as they have 67-256 classes
and about 2-10 times the image size than earlier at about 8K, 15K
and 30K. We set the initial truncation level to about twice the ground
truth at 200, 100 and 500 respectively. For batch learners such as
MM or VI, they cannot handle these kind of dataset as the entire datset
size is too large. Instead, we fix the batch size at around 5K and
take a random subset of the entire dataset that do not change for
each iteration. Meaning there is a large portion of dataset that MM
will never see. This is where SGAs are handy as it can skip and access
the entire dataset when the number of iterations increases as compared
to MM. We set the minibatch size to 1515 (10 random samples per class),
1005 (15 random samples per class) and 2560 (10 random samples per
class) respectively. We can see that after 30 iterations, SGAs outperforms
MM once again on all 5 criteria.

Lastly, we look at dataset \#6 (SUN397). This dataset is considered
quite large at >0.1M images and almost 400 class. Using 1000 classes
for truncation level, we see that MM is unable to learn as quickly
as SGAs after 30 iterations as the estimated model is quite large
at 838 compared to 570.5 for the SGA+F. Also for SGAs, there are some
minor improvements for NMI and ACC as compared to MM. The computational
time is also 30\% faster for SGA+F.

\begin{figure}
\begin{centering}
\includegraphics[scale=0.4]{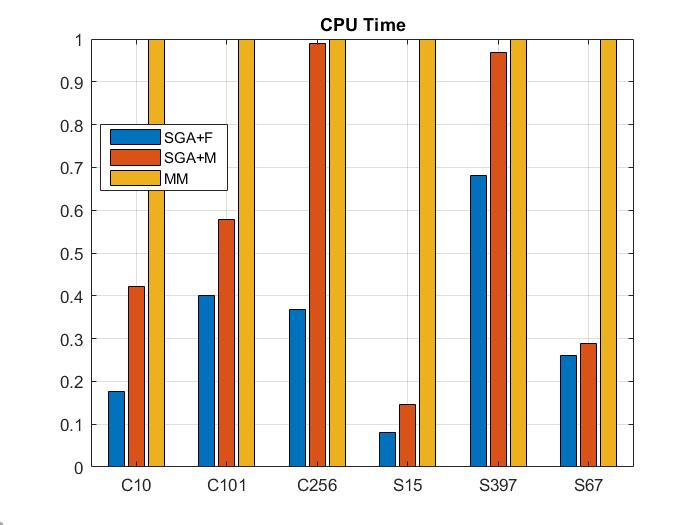}
\par\end{centering}
$\;$

$\;$

$\;$
\begin{centering}
\includegraphics[scale=0.4]{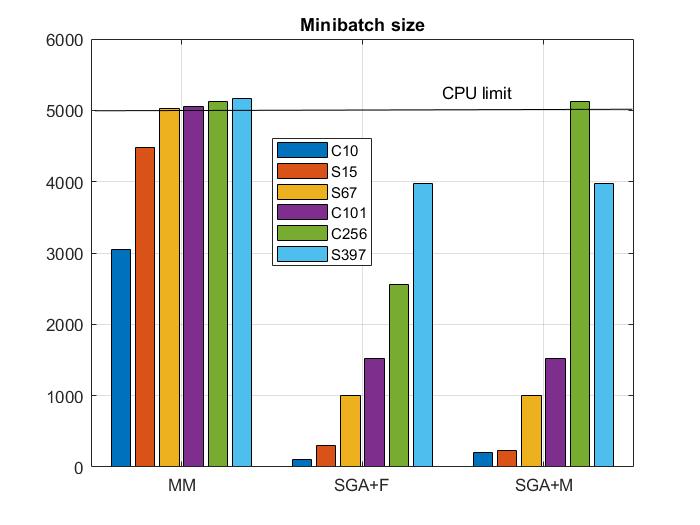}
\par\end{centering}
\caption{CPU time (as a factor of MM) and minibatch size}
\end{figure}

\begin{figure}
\begin{centering}
\includegraphics[scale=0.4]{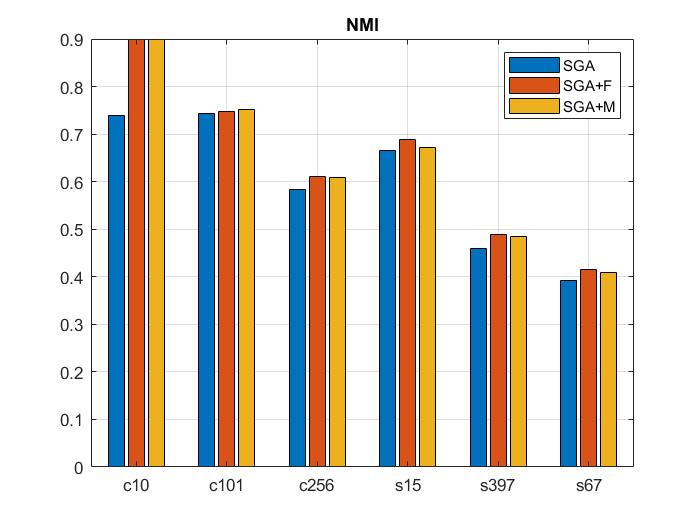}
\par\end{centering}
$\;$

$\;$

$\;$
\begin{centering}
\includegraphics[scale=0.4]{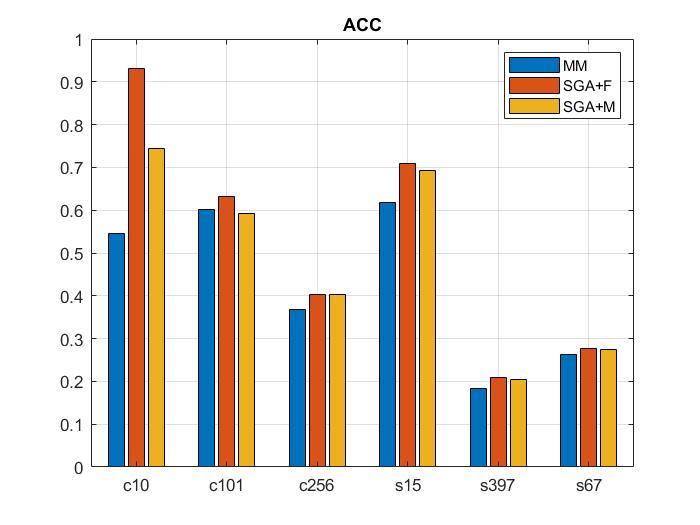}
\par\end{centering}
\caption{NMI and Accuracy}
\end{figure}

\begin{table}
\begin{centering}
\begin{tabular}{|c||c||c||c||c|}
\hline 
Gnd. Truth & Trunc. Lv. & MM & SGA+F & SGA+M\tabularnewline
\hline 
\hline 
10 & 50 & 13.3 & \textbf{11.7} & 24\tabularnewline
\hline 
\hline 
101 & 200 & 126.3 & \textbf{118.7} & 139.7\tabularnewline
\hline 
\hline 
256 & 500 & 464.7 & \textbf{314.2} & 392.7\tabularnewline
\hline 
\hline 
15 & 50 & \textbf{17.7} & 18 & 19\tabularnewline
\hline 
\hline 
67 & 100 & 81.8 & \textbf{74.6} & 84.3\tabularnewline
\hline 
\hline 
397 & 1000 & 838 & \textbf{570.5} & 577\tabularnewline
\hline 
\end{tabular} 
\par\end{centering}
\caption{Model Selection}
\end{table}

\section{Experiment 2: Empirical convergence analysis of SGA}

In this section, we compute the convergence plot when learning (SGA
or SGA+Fisher) the DPM model on the Caltech256 and SUN397 datasets
in Fig. 3 and 4 by computing the negative lower bound. 

In the top rows, we have 2 graphs for SGA and SGA+Fisher computed
on Caltech256 for a computational budget of 35 iterations. We observe
that both methods are not able to converge after 35 iterations. However,
SGA+Fisher is able to peak faster than SGA and exhibits less fluctuation
after 20 iterations. 

In the bottom row, we perform the same comparison task on SUN397.
We observe that both SGA and SGA+Fisher converges after around 30
iterations but the curve of SGA+Fisher increases much steeper in the
early few iterations.

In conclusion, due to constant stepsize, SGA exhibit more fluctuation.
However, we see that for SGA+Fisher, the use of Fisher information
can reduce this fluctuation as well as enjoy a faster learning curve.

\begin{figure}
\begin{centering}
\includegraphics[scale=0.6]{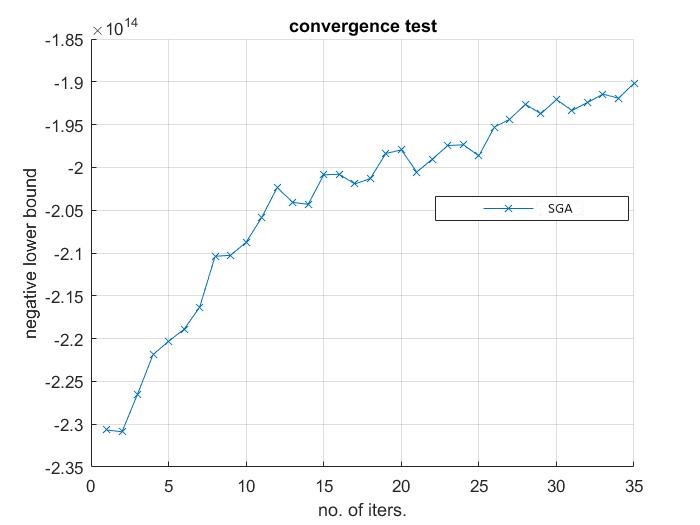}
\par\end{centering}
$\;$

$\;$

$\;$
\begin{centering}
\includegraphics[scale=0.6]{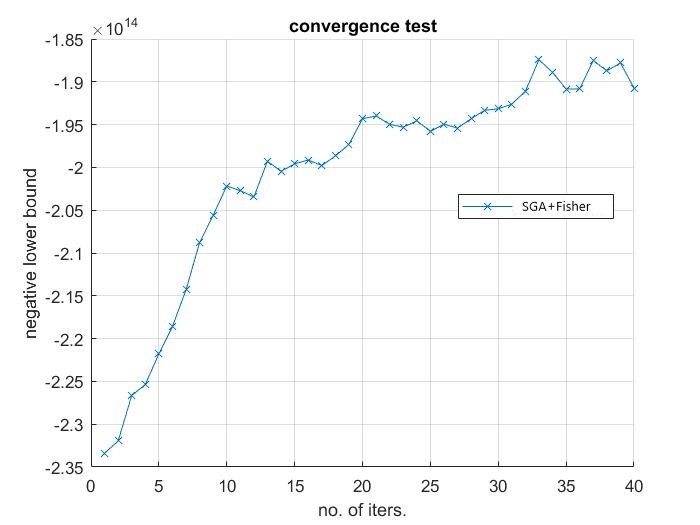}
\par\end{centering}
\caption{Convergence plots of SGA (top) and SGA+Fisher (bot) on Caltech256 }
\end{figure}

\begin{figure}
\begin{centering}
\includegraphics[scale=0.6]{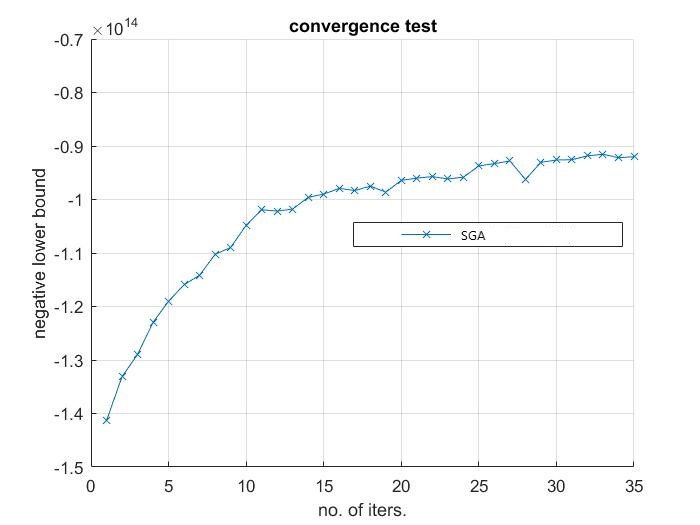}
\par\end{centering}
$\;$

$\;$

$\;$
\begin{centering}
\includegraphics[scale=0.6]{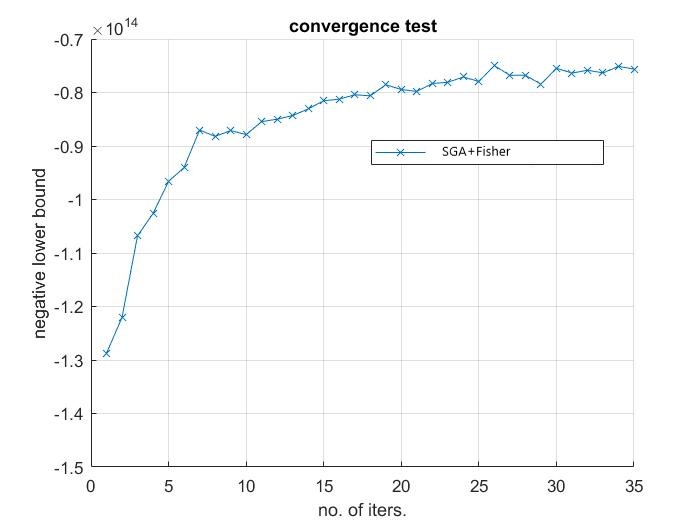}
\par\end{centering}
\caption{Convergence plots of SGA (top) and SGA+Fisher (bot) on SUN397}
\end{figure}

\begin{table}
\begin{centering}
\begin{tabular}{|c|c|c|c|c|}
\cline{2-5} \cline{3-5} \cline{4-5} \cline{5-5} 
\multicolumn{1}{c|}{} & 10 & 15 & 67 & 397\tabularnewline
\hline 
MMGM \cite{chen2014robust}; & - & 0.186 & - & -\tabularnewline
 &  & - &  & \tabularnewline
\hline 
DDPM-L \cite{nguyen2017discriminative}: & - & 0.218 & - & -\tabularnewline
 &  & - &  & \tabularnewline
\hline 
DPmeans \cite{Kulis2012}: & 0.418 & 0.316 &  & \tabularnewline
 & - & - &  & \tabularnewline
\hline 
VB-DPM \cite{Blei2006}: & 0.446 & 0.317 & - & -\tabularnewline
 & - & - &  & \tabularnewline
\hline 
MM-GMM \cite{lim2018fast}: & 0.386 & 0.310 & - & -\tabularnewline
 & <40.73> & <33.58> &  & \tabularnewline
\hline 
MM-DPM \cite{lim2018fast}: & 0.387 & 0.321 & - & -\tabularnewline
 & <39.14> & <33.59> &  & \tabularnewline
\hline 
Kmeans \cite{wang2017unsupervised}: & - & 0.659 & 0.386 & -\tabularnewline
 &  & <65.0> & <35.6> & \tabularnewline
\hline 
LDPO-A-FC \cite{wang2017unsupervised}: & - & \textbf{0.705} & 0.389 & -\tabularnewline
 &  & \textbf{<73.1>} & \textbf{<37.9>} & \tabularnewline
\hline 
OnHGD \cite{fan2016online}: & - & - & - & -\tabularnewline
 &  & <67.34> &  & <26.52>\tabularnewline
\hline 
SVI \cite{Hoffman2013} & 0.810 & 0.670 & 0.413 & 0.489\tabularnewline
 & <78.98> & <63.56> & <30.76> & <23.93>\tabularnewline
\hline 
SGA+Fisher (ours) & \textbf{0.898} & \textbf{0.70} & \textbf{0.53} & \textbf{0.60}\tabularnewline
 & \textbf{<93.1>} & <70.3> & \textbf{<37.8>} & \textbf{<31.9>}\tabularnewline
\hline 
\end{tabular}
\par\end{centering}
\caption{Recent Bayesian nonparametrics vs best proposed result on the object
and scene categorization dataset. Evaluation using NMI and Accuracy
(in diamond bracket).}
\end{table}

\section{Comparison with literature}

In this section, we are mainly comparing the clustering performance
of our proposed method with results reported by the state\textendash of-the-arts
methods.

\subsection{Baselines}

We conducted a literature search of recent Bayesian nonparametrics
citing the datasets we use. We first discuss the baselines in Table
3 as follows: MMGM \cite{chen2014robust}, DDPM-L \cite{nguyen2017discriminative},
LDPO \cite{wang2017unsupervised}, DPmeans \cite{Kulis2012}, VB-DPM
\cite{Blei2006}, MM-GMM, MM-DPM \cite{lim2018fast} and OnHGD\cite{fan2016online}.
We also implemented SVI \cite{Hoffman2013} for DPM. 

\subsection{Features}

MMGM and DDPM-L are using the 128 dimensional SIFT features. DPmeans,
MM-GMM, VB-DPM and MM-DPM are using the 2048 dimensional sparse coding
based Fisher vector in \cite{lim2018fast}. For LDPO, the authors
use the FC7 of Alexnet pretrained on Imagenet. We did not cite the
patch mining variant of LDPO as it is mainly related to feature extraction
rather than BNP, hence it is not a direct comparison. For our implementation
of SVI, we use VGG16 pretrained on ImageNet. For our proposed method,
we use VGG16 pretrained on Place205 as the image feature extractor.

\subsection{Truncation level}

The truncation level of MMGM and DDPM-L are not given in \cite{nguyen2017discriminative}.
For LDPO, it is fixed to the ground truth \cite{wang2017unsupervised}.
For MM-GMM, VB-DPM, MM-DPM and SVI, the truncation setting are identical
to this work. For DPmeans, we use ground truth to initialize its parameter
for convenience as reported in \cite{lim2018fast}.

\subsection{Comparisons on Caltech10}

We find that none of the baselines can outperform the proposed method
partly due to the discriminative power that deep feature offers and
the use of the proposed stochastic optimization approach.

\subsection{Comparisons on Scene15 \& MIT67}

In Table 6, LDPO-A-FC performs better than SGA+Fisher on Accuracy
for Scene15 but underperformed for NMI on MIT67. In fact, LDPO is
only on par with Kmeans for NMI on MIT67.

For DDPM-L and MMGM, there is a huge disadvantage on their results
due to using handcrafted feature. However, from their extended results
in \cite{nguyen2017discriminative} we infer that the performance
of DDPM-L should be slightly better than DPM using MM when the same
deep feature is used. 

In \cite{lim2018fast}, MM-DPM is shown to outperforms its VB-DPM
counterpart on scene15. Given that MM (of DPM) when assuming $d=0$
is almost similar to MM-DPM except without modeling precision, the
result of MM-DPM should be slightly better than MM (of DPM) when using
deep features. 

\subsection{Comparisons on Caltech101, Caltech256 and SUN397}

To the best of our knowledge, it is very rare to find any recent BNP
works addressing datasets beyond 67 classes for image datasets. The
main reason is that it is difficult to scale up BNP, which is one
of the main reason why we proposed the SGA learner over MM. 

Although, the authors in \cite{fan2016online} applied a BNP model
to SUN397, they mainly use it for learning a Bag-of-Words representation
(also discussed in their batch learning counterpart \cite{fan2016variational}).
It appears they then use a supervised learner such as Bayes's decision
rule for classification as suggested in \cite{fan2016variational}.
No model selection was mentioned for SUN397 either. In \cite{fan2016variational},
the authors made a comparison using variational inference on Dirichlet
process with Gaussian mixture (DP-GMM) and with generalized Dirichlet
mixture (DP-GD). Their results showed that DP-GD has 4\% Accuracy
gain over DP-GM on the Cats and Dogs dataset. Consequently, they also
showed that using a hierarchical Dirichlet process (HDP) further improved
their results by 1-2\% on Accuracy. For SUN397, the Accuracy reported
in \cite{fan2016online} for their online hierarchical Dirichlet process
mixture of generalized Dirichlet (onHGD) is 26.52\% on SUN397. It
is somewhat comparable to our Accuracy of 20.9\% using SGA since we
do not use hierarchical representation nor generalized Dirichlet mixture
(combined of both gives 5-6\% gain according to \cite{fan2016variational}).
The authors of \cite{fan2016online} also reported an Accuracy of
67.34\% for SUN16. Although this is not a direct comparison to Scene15
as the classes in SUN16 are different, it provides some insight on
our method. Our only concern is whether their result is attainable
using SIFT. Once again, we should point out that in \cite{fan2016online}
as also mentioned in \cite{fan2016variational}, the authors use OnHGau
and OnHGD for learning a bag-of-words representation while our method
mainly learns the cluster mean of each class in Table 3.

\section{Conclusion}

Scalable algorithms of variational inference allows Bayesian nonparametrics
such as Dirichlet process mixture to scale up to larger dataset at
fractional cost. In this paper, we target up to about 100K images
and 400 object classes and high dimensional features using VGG16 pretrained
on ImageNet. However, the main problem is that most scalable Dirichlet
process mixture algorithms today still rely on the closed form learning
found in variational inference from the past two decades. Stochastic
gradient ascent is a modern approach to machine learning and it is
widely deployed in the training of deep neural networks. However,
variational inference and stochastic gradient ascent are mutually
exclusive. Recently, stochastic gradient ascent with constant stepsize
has been discussed for variational inference, In this work, we propose
using stochastic gradient ascent as a learner in variational inference.
Unlike the latter, stochastic gradient ascent do not require closed
form expression for the variational posterior expectatations. However,
stochastic gradient ascent alone is not optimal for learning. It suffers
from being a local optimum approach and when applied to stochastic
dataset, it takes a long time to converge. Stochastic optimization
techniques refine stepsizes for guaranteed convergence and better
local search. Specifically, we explored Fisher information and the
momentum method to achieve a faster convergence. We compare our new
stochastic learner on the Dirichlet process Gaussian mixture. We showed
the performance gained in terms of NMI, accuracy, model selection
and computational time on large scale datasets such as the Caltech101,
Caltech256 and SUN397.

\section{Reference}

\bibliographystyle{elsarticle-num}
\addcontentsline{toc}{section}{\refname}\bibliography{allmyref}

\end{document}